%% file: main.tex
\documentclass[10pt,twocolumn,letterpaper]{article}

\usepackage{cvpr}              

\usepackage{graphicx}
\usepackage{amsmath}
\usepackage{amssymb}
\usepackage{booktabs}
\usepackage{bbding}
\usepackage{graphicx}
\usepackage{booktabs}
\usepackage{cite}
\usepackage{color}
\usepackage{xspace}
\usepackage{ctable}
\usepackage{colortbl}
\usepackage{subcaption}
\usepackage{multirow}
\usepackage{float}
\usepackage{stfloats}
\usepackage{stmaryrd}
\usepackage{algorithm}
\usepackage{algorithmicx}
\usepackage{algpseudocode}
\usepackage{threeparttable}
\usepackage{wrapfig}
\usepackage{pifont}
\usepackage{caption}[font={small}]
\usepackage{wrapfig}
\usepackage[pagebackref,breaklinks,colorlinks]{hyperref}

\usepackage[pagebackref,breaklinks,colorlinks]{hyperref}

\usepackage[capitalize]{cleveref}
\crefname{section}{Sec.}{Secs.}
\Crefname{section}{Section}{Sections}
\Crefname{table}{Table}{Tables}
\crefname{table}{Tab.}{Tabs.}

\def\cvprPaperID{5325} 
\def\confName{CVPR}
\def\confYear{2023}

\begin{document}

\title{LOGO: A Long-Form Video Dataset for Group Action Quality Assessment}
\author{
    Shiyi Zhang\textsuperscript{1,2,3},
    Wenxun Dai\textsuperscript{1},
    Sujia Wang\textsuperscript{1},    
    Xiangwei Shen\textsuperscript{1},
    Jiwen Lu\textsuperscript{2,3},
    Jie Zhou\textsuperscript{2,3},
    Yansong Tang\textsuperscript{1,}\thanks{\;indicates the corresponding author.}\\
    \textsuperscript{1} Shenzhen International Graduate School, Tsinghua University\\
    \textsuperscript{2} Department of Automation, Tsinghua University\\
    \textsuperscript{3} Beijing National Research Center for Information Science and Technology\\
    {\tt \small \{shiyi-zh19@mails.,lujiwen@,jzhou@,tang.yansong@sz.\}tsinghua.edu.cn}\\
}


\makeatletter
\let\@oldmaketitle\@maketitle
\renewcommand{\@maketitle}{\@oldmaketitle
\begin{minipage}{\textwidth}
\vspace{-0.6cm}
\centering
\includegraphics[width=\linewidth]{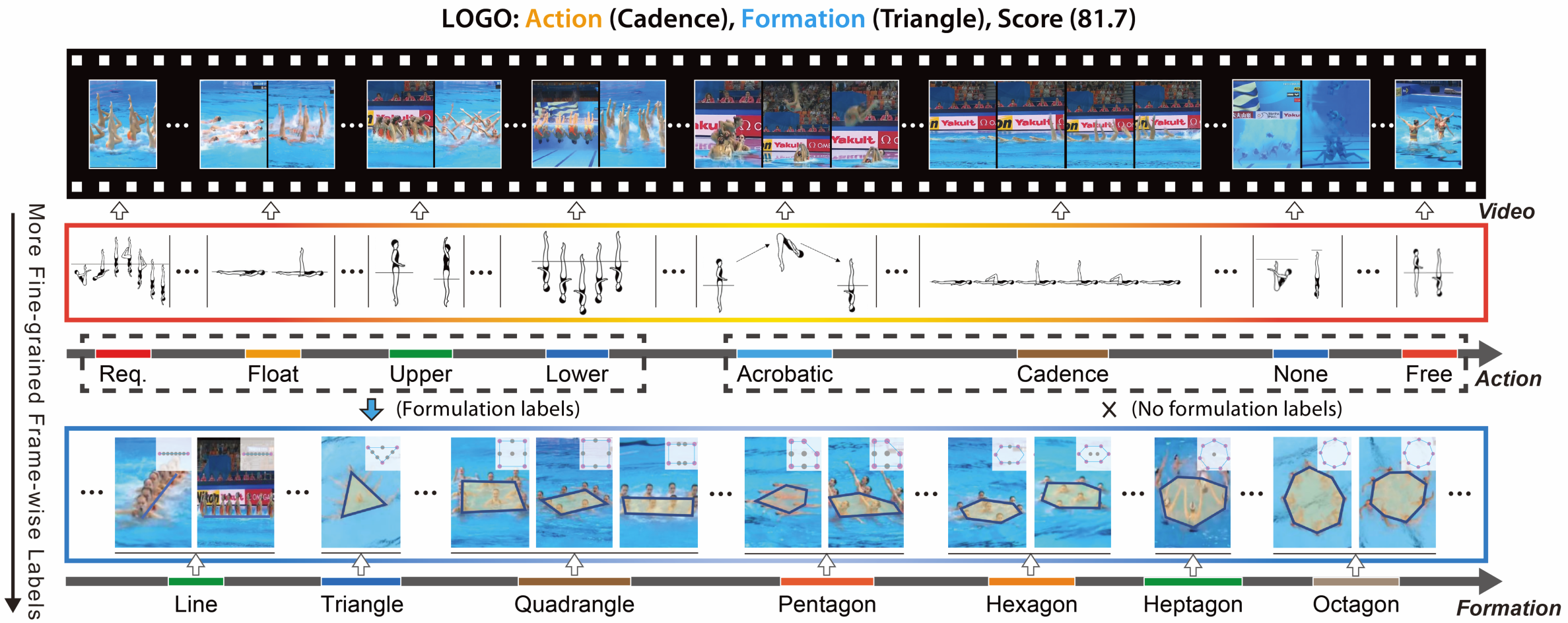}
    \captionof{figure}{An overview of the \textbf{\textit{LOGO}} dataset. LOGO is a multi-person long-form video dataset with frame-wise annotations on both action procedures (as shown in the second line) and formations (as shown in the third line, which reflects relations among actors) based on artistic swimming scenarios. It provides a potential for constructing an action quality assessment approach with the ability to model group information among actors. Longer video durations also challenge the ability of the method to aggregate long-term temporal information.}
    \label{fig:datasetoverview}
    \vspace{+0.6cm}
\end{minipage}}
\makeatother

\maketitle
\input{sections/0_abstract}
\input{sections/1_introduction}
\input{sections/2_relatedwork}
\input{sections/3_thelogodataset}

\input{sections/4_approach}
\input{sections/5_experiments}

\input{sections/6_conclusion}

{\small
\bibliographystyle{ieee_fullname}
\bibliography{references}
}

\end{document}

%% file: sections/0_abstract.tex
\begin{abstract}
Action quality assessment (AQA) has become an emerging topic since it can be extensively applied in numerous scenarios. However, most existing methods and datasets focus on single-person short-sequence scenes, hindering the application of AQA in more complex situations. To address this issue, we construct a new multi-person long-form video dataset for action quality assessment named LOGO. Distinguished in scenario complexity, our dataset contains 200 videos from 26 artistic swimming events with 8 athletes in each sample along with an average duration of 204.2 seconds. As for richness in annotations, LOGO includes formation labels to depict group information of multiple athletes and detailed annotations on action procedures. Furthermore, we propose a simple yet effective method to model relations among athletes and reason about the potential temporal logic in long-form videos. Specifically, we design a group-aware attention module, which can be easily plugged into existing AQA methods, to enrich the clip-wise representations based on contextual group information. To benchmark LOGO, we systematically conduct investigations on the performance of several popular methods in AQA and action segmentation. The results reveal the challenges our dataset brings. Extensive experiments also show that our approach achieves state-of-the-art on the LOGO dataset. The dataset and code will be released at \url{https://github.com/shiyi-zh0408/LOGO}.
\end{abstract}

%% file: sections/1_introduction.tex
\section{Introduction}
\input{tables/dataset_comparison}

Action quality assessment (AQA) is applicable to many real-world contexts where people evaluate how well a specific action is performed such as sports events\cite{gordon1995automated, parmar2019action, jug2003trajectory, pervse2007automatic, pirsiavash2014assessing, parmar2019and, venkataraman2015dynamical, parmar2017learning}, healthcare\cite{malpani2014pairwise, zhang2014relative, sharma2014video, zhang2011video, zia2015automated, zia2018video}, art performances, military parades, and others. Due to the extensive application of AQA, many efforts have been made over the past few years. Although some existing works have achieved promising performances in several simple scenarios, the application of AQA in many situations is still difficult to implement. In the data-driven era, the richness of the dataset largely determines whether the model can be applied to a wide range of scenarios or not. Inspired by this, we reviewed the existing datasets in AQA and concluded that they are not rich enough for the following two reasons:

\textbf{Simplicity of Scenarios.}  We argue that the complexity of the AQA application scenes is reflected in two aspects, the number of people and the duration of the videos. In snowboarding, for example, there is only one performer, while there are multiple actors in a military parade. In diving, the duration is only about 5 seconds, while in artistic swimming and dance performances, the duration of the action reaches several minutes. However, most existing datasets contain a single performer in each sample and many of them collect videos of 3-8s\cite{parmar2019and, parmar2019action, xu2022finediving, parmar2017learning, pirsiavash2014assessing}, which makes it difficult for existing methods to model complex scenes with more actors and longer duration. In such cases, just focusing on how well actions are performed by each individual may be insufficient. Relations among actors should be built, and the potential temporal logic in long-term videos should be modeled.

\textbf{Coarse-grained Annotations.} Though there have been some long-form video datasets in AQA\cite{zeng2020hybrid,xu2019learning} that provide more complex scenes, they typically contain the score as the only annotation. Such coarse-grained annotation makes it difficult for models to learn deeper information, especially in more complex situations. Simply judging action quality via regressing a score for a long-term video could be confusing since we cannot figure out how the model determines whether an action is well-performed or not. 

To address these issues, we propose a multi-person long-form video dataset, LOGO (short for LOng-form GrOup). With 8 actors in each sample, LOGO contains videos with an average duration of 204.2s, much longer than most existing datasets in AQA, making the scenes more complex. Besides, as shown in Figure \ref{fig:datasetoverview},  LOGO contains fine-grained annotations including frame-wise action type labels and temporal boundaries of actions for each video. We also devise formation labels to depict relations among actors. Specifically, we use a convex polygon to represent the formation actors perform, which reflects their position information and group information. In general, LOGO is distinguished by its scenario complexity, while it also provides richer annotations compared to most existing datasets in AQA\cite{parmar2019and, parmar2019action, xu2022finediving, parmar2017learning, pirsiavash2014assessing,zeng2020hybrid,xu2019learning}. 

Furthermore, we build a plug-and-play module, GOAT (short for GrOup-aware ATtention), to bridge the gap between single-person short-sequence scenarios and multi-person long-sequence scenarios. Specifically, in the spatial domain, by building a graph for actors, we model the relations among them. The nodes of this graph represent actors’ features extracted from a CNN, and the edges represent the relations among actors. Then we use a graph convolution network (GCN)\cite{kipf2016semi} to model the group features from the graph. The optimized features of the graph then serve as ``\textit{queries}" and ``\textit{keys}" for GOAT. In the temporal domain, after feature extraction by the video feature backbone, the clip-wise features serve as ``\textit{values}" for GOAT. Instead of fusing the features simply using the average pooling as most previous works\cite{tang2020uncertainty, yu2021group, xu2022finediving}, GOAT learns the relations among clips and models the temporal features of the long-term videos based on the spatial information in every clip. 

The contributions of this paper can be summarized as: (1) We construct the first multi-person long-form video dataset, LOGO, for action quality assessment. To the best of our knowledge, LOGO stands out for its longer average duration, the larger number of people, and richer annotations when compared to most existing datasets. Experimental results also reveal the challenges our proposed dataset brings. (2) We propose a plug-and-play group-aware module, GOAT, which models the group information and the temporal contextual relations for input videos, bridging the gap between single-person short-sequence scenarios and multi-person long-sequence scenarios. (3) Experimental results demonstrate that our group-aware approach obtains substantial improvements compared to existing methods in AQA and achieves the state-of-the-art.

%% file: tables/dataset_comparison.tex

\begin{table*}[t]
    \centering
    \vspace{2mm}
    \caption{Comparisons of LOGO and existing datasets of action quality assessment (upper part of the table) and group activity recognition (lower part of the table). \textit{Score} indicates the score annotations; \textit{Action} denotes action types and temporal boundaries; \textit{Act.Label} indicates the action types of both individuals and groups. \textit{Bbox} indicates bounding boxes for actors. \textit{Formation} represents formation annotations. \textit{Temp.} indicates temporal boundary, \textit{Spat.} indicates spatial localization. }
    \label{tab:action_set_comparison}
    \setlength{\tabcolsep}{1.5mm}{
    \begin{tabular}{lccccccc}
    \toprule[1.6pt]
        \textbf{Dataset} & \textbf{Duration} & \textbf{Avg.Dur.} & \textbf{Anno.Type} & \textbf{Samples} & \textbf{Events} & \textbf{Form.Anno.} &\textbf{Year}\\ 
        \midrule
        MTL Dive\cite{pirsiavash2014assessing} & 15m,54s & 6.0s & Score & 159 & 1 & \ding{55} & 2014\\ 
        UNLV Dive\cite{parmar2017learning} & 23m,26s & 3.8s & Score & 370 & 1 & \ding{55} & 2017\\
        AQA-7-Dive\cite{parmar2019action} & 37m,31s & 4.1s & Score & 549 & 6 & \ding{55} & 2019\\
        MTL-AQA\cite{parmar2019and} & 96m,29s & 4.1s & Action,Score & 1412 & 16 & \ding{55} & 2019\\
        Rhythmic Gymnastics\cite{zeng2020hybrid} & 26h,23m,20s & 1m,35s & Score & 1000 & 4 & \ding{55} & 2020\\
        FSD-10\cite{liu2020fsd} & - & 3-30s & Action,Score & 1484 & - & \ding{55} & 2020\\
        FineDiving\cite{xu2022finediving} & 3h,30m,0s & 4.2s & Action,Score & 3000 & 30 & \ding{55} & 2022\\
        
        \midrule
        Collective Activity\cite{choi2009they} & - & - & Act.Label & 44 & - & \ding{55} & 2009\\
        NCAA\cite{ramanathan2016detecting} & 16h,9m,52s & 4s & Bbox,Act.Label & 11436 & - & \ding{55} & 2016\\
        Volleyball\cite{ibrahim2016hierarchical} & 2h,12m,1s & 1.64s & Bbox,Act.Label & 4830 & - & \ding{55} & 2016\\
        FineGym\cite{shao2020finegym} & 161h,1m,45s & 45.7s & Act.Label,Temp.& 12685 & 10 & \ding{55} & 2020\\
        Multisports\cite{li2021multisports} & 18h,34m,40s & 20.9s & Act.Label,Temp.,Spat.& 3200 & 247& \ding{55} & 2021\\
    \midrule
        \textbf{LOGO}(Ours) & 11h,20m,41s & 3m,24s & Action,Formation,Score & 200 & 26 & \ding{51}(15764) &\\
    \bottomrule[1.6pt]
    \end{tabular}}
\end{table*}

%% file: sections/2_relatedwork.tex
\section{Related Work}
\textbf{Action Quality Assessment.} Assessing action quality is an increasingly popular trend in computer vision with wide applications such as video retrieval, instructional video analysis, \textit{etc}. A wide variety of methods for formulating AQA tasks have been proposed\cite{pirsiavash2014assessing, gordon1995automated, parmar2017learning, tang2020uncertainty, li2018scoringnet, yu2021group, bertasius2017baller, doughty2019pros, xu2019learning, zeng2020hybrid, xu2022likert}. In long video AQA tasks, actions performed at different times may significantly impact the scores given by experts. However, only a handful of works directly address the problem of long-term video AQA\cite{bertasius2017baller, doughty2019pros, xu2019learning, zeng2020hybrid, xu2022likert}. There are many datasets for AQA, including Diving\cite{pirsiavash2014assessing,parmar2017learning, xu2022finediving, parmar2019action, parmar2019and}, Figure Skating\cite{pirsiavash2014assessing, liu2020fsd}, Gymnastic Vault\cite{parmar2017learning, parmar2019action}, Basketball\cite{bertasius2017baller}, Fitness\cite{tang2022flag3d} and Rhythmic Gymnastics\cite{zeng2020hybrid}. However, the problem of considering group information in multi-person sports has been relatively unexplored. As shown in Table \ref{tab:action_set_comparison}, MIT-Dive\cite{pirsiavash2014assessing}, UNLV-Dive\cite{parmar2017learning}, AQA-7-Dive\cite{parmar2019action} and Rhythmic Gymnastics\cite{zeng2020hybrid} only provide action scores, while MTL-AQA\cite{parmar2019and} FSD-10\cite{liu2020fsd} and FineDiving\cite{xu2022finediving} only provide action types and scores. To perform AQA in long videos with multi-person, we construct a long-form artistic swimming video dataset, LOGO, with fine-grained frame-wise action and formation labels. Furthermore, we propose a method to model relations among athletes and reason about the potential temporal logic in long-form videos.

\textbf{Group Activity Understanding.} 
Group activity understanding, which concentrates on interpreting the collective activity from multi-person behavioral and interaction dynamics, has attracted plenty of work recently. The focus of existing methods has shifted from using shallow hand-crafted features \cite{lan2011discriminative, lan2012social, ramanathan2013social, choi2012unified, choi2011learning, shu2015joint} to using deep neural networks\cite{ibrahim2018hierarchical, qi2018stagnet, gavrilyuk2020actor, li2021groupformer, bagautdinov2017social, thilakarathne2021pose}. Diverse datasets have been proposed to facilitate research in this area, such as the Collective Activity dataset \cite{choi2009they}, Volleyball dataset \cite{ibrahim2016hierarchical}, etc. However, the methods and datasets mentioned above only stay at the level of group activity recognition; a more profound understanding of group activities should be able to not only find the high-level group activity but also evaluate the quality of group activities, which requires more attention to the quality of collaboration among actors and fine-grained actions of actors. In this paper, LOGO takes the first step toward quality assessment and sets a new benchmark for group activity understanding.

%% file: sections/3_thelogodataset.tex
\input{figures/formation.tex}
\section{The LOGO Dataset}
\label{sec:thelogodataset}

We propose a new multi-person long-form video dataset with detailed annotations on action and formation, LOGO, to set a new challenging benchmark for AQA. We will introduce it from its construction and annotation in this section.
\subsection{Dataset Construction}
\textbf{Collection.} According to the official FINA rules, the same rules, and prescribed movements are used from 2018 until the end of the 2022 World Championships. So we collected the officially designated \textit{Technical} and \textit{Free} artistic swimming competitions during this period (2 World Championships, 1 Olympic Games, and 4 World Series). We download competition videos with high resolution, e.g., 720p and 1080p, on online video platforms like YouTube. Then we sampled them at 25fps and 1fps sampling rates for fine-grained action and formation labeling. Each official videos provide various content, including three sub-scores, one final score, and frames from different views.

\textbf{Lexicon.} We construct a fine-grained video dataset organized by temporal structure, which contains action and formation manual annotations, shown in Figure \ref{fig:formations}. Herein, we design the labeling system with professional artistic swimming athletes to construct a lexicon for annotation, considering FINA rules and the actual scenario of the competitions. In the \textit{Technical} event, the group size is eight people, the video length is 170±15s, and the actions include \textit{Upper}, \textit{Lower}, \textit{Float}, \textit{None}, \textit{Acrobatic}, \textit{Cadence}, and five    \textit{Required Elements}. Each competition cycle needs to complete five \textit{Required Elements}, at least two \textit{Acrobatic} movements, and at least one \textit{Cadence} action. In the \textit{Free} events, there are 8 people, the video length is 240±15s, and the actions include \textit{Upper}, \textit{Lower}, \textit{Float}, \textit{None}, \textit{Acrobatic}, \textit{Cadence}, and \textit{Free} elements. When performing \textit{Required}, \textit{Upper}, \textit{Lower}, and \textit{Float}, the athletes form neat polygons as in Figure \ref{fig:formations}.

\input{figures/dataset_statics}

\textbf{Annotation.} Given an RGB artistic swimming video, the annotator utilizes our defined lexicon to label each frame with its action and formation. We accomplish the 25fps frame-wise action annotation stage utilizing the COIN annotation toolbox\cite{tang2019coin} and the 1fps frame-wise formation labels using Labelme. Specifically, we set strict rules defining the boundaries between artistic swimming sequences and the formation marking position and employ eight workers with prior knowledge in the artistic swimming domain to label the dataset frame by frame following the rules. The annotation results of one worker are checked and adjusted by another, which ensures annotation results are double-checked. Under this pipeline, the total time for the annotation process is above 600 hours.

\subsection{Dataset Statics}
The LOGO dataset consists of 200 video samples from 26 events with 204.2s average duration and above 11h total duration, covering 3 annotation types, 12 action types, and 17 formation types.
Figure \ref{fig:datasetstatics} shows the score and action type distributions among all the events. Table \ref{tab:action_set_comparison} reports more detailed information on the LOGO dataset and compares it with existing AQA datasets and other fine-grained sports datasets. Our dataset differs from existing AQA datasets in the annotation type and dataset scale. Specifically, our dataset includes formation types providing group information and longer videos on sports, helping to evaluate the quality of actions in multiplayer sports quantitatively. Other fine-grained sports datasets cannot be used for assessing action quality due to a lack of action scores. LOGO is the first fine-grained sports video dataset for Group AQA, filling the fine-grained group annotations void in AQA.


%% file: figures/formation.tex
\begin{figure}[t]
    \centering
    \includegraphics[width=1\linewidth]{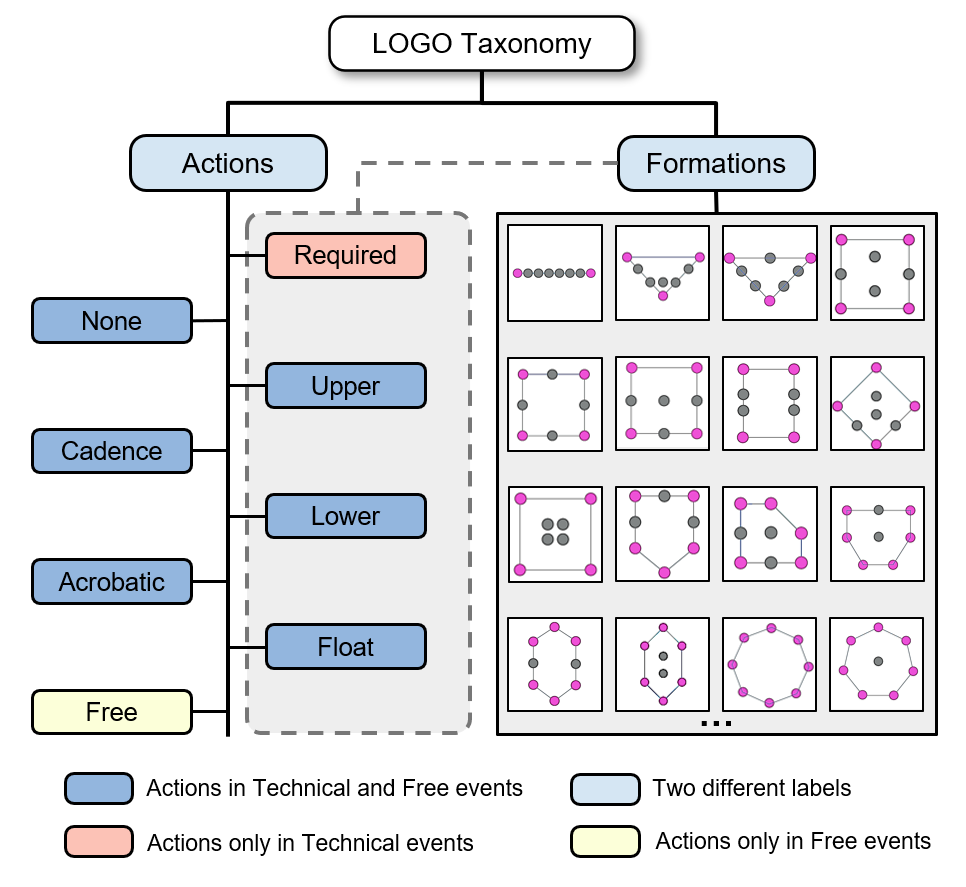}
    \caption{A tree structure of LOGO Taxonomy. LOGO organizes both the \textit{Actions} and \textit{Formations} annotations hierarchically. The left part shows the \textit{Actions} categories of \textit{Technical} and \textit{Free} events. The right part depicts the formation annotation instances when the group is doing \textit{Required}, \textit{Upper}, \textit{Lower} or \textit{Float} actions (the right sub-tree of \textit{Actions}) and not when the group is doing other actions, during which the formations are indistinguishable.
}
    \label{fig:formations}
    \vspace{-12pt}
\end{figure}

%% file: figures/dataset_statics.tex
\begin{figure}[t]
    \centering
    \includegraphics[width=1\linewidth]{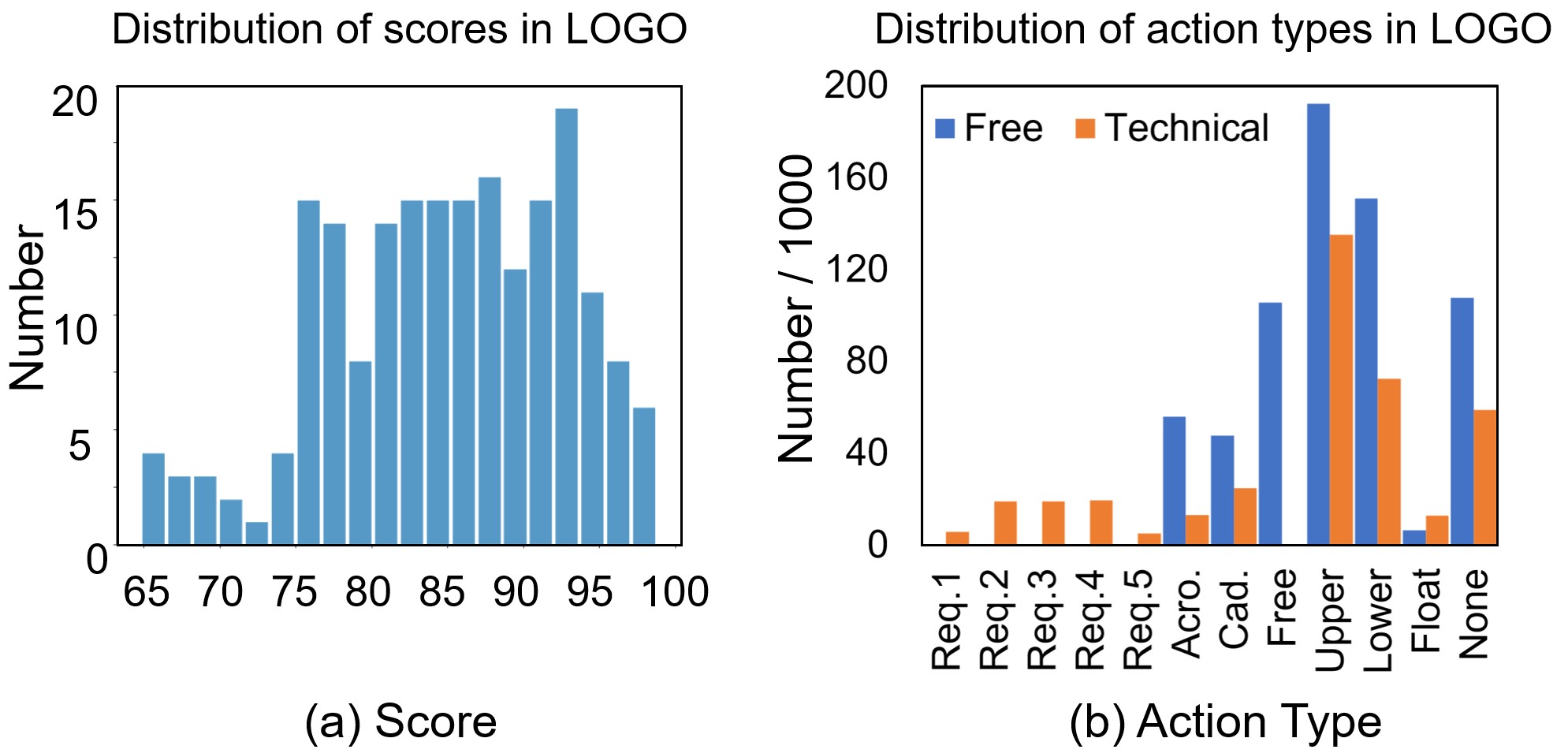}
    \caption{Statics of LOGO. (a) The score distribution of videos. (b) The action-type distribution of frames.}
    \label{fig:datasetstatics}
    \vspace{-12pt}
\end{figure}

%% file: sections/4_approach.tex
\section{Approach}
\label{sec:approach}
\input{figures/goat}


\subsection{Problem Formulation}
\label{subsec:problemformulation}
Most existing methods in AQA process the output of the video feature backbone using average pooling before regression. It can be represented as:
\begin{equation}
    \hat{y}=\mathcal{R}[\frac{1}{T}\sum_{i=1}^T\mathcal{F}(X_i)|\Theta],
\end{equation}where $X_i$ denotes the $i-$th clip of the video. $\mathcal{F}$ indicates the video feature backbone; $T$ is the number of clips; $\mathcal{R}$ represents the regression algorithm; $\Theta$ is the learnable parameters of $\mathcal{R}$ and $\hat{y}$ is the predicted score.
However, fusing temporal information simply by average pooling in scenes with longer duration may be confusing since there could be redundant or even invalid information in the video. Besides, when dealing with multi-person scenarios, we need to depict the relations among actors. We use group information to fuse the extracted features in the temporal axis, modeling the relations among clips. It can be represented as:
\begin{gather}
    \hat{y}=\mathcal{R}[\frac{1}{T}\sum_{i=1}^Tf_i|\Theta],\\
    f_i=\sum_{j=1}^T\mathcal{G}(X_i,X_j)\mathcal{F}(X_j),  i=1...T,
    \label{con:fuse}
\end{gather}where $\mathcal{G}(X_i,X_j)$ denotes the weight computed by group features of the $i$-th clip and the $j$-th clip; $f_i$ indicates the reconstructed feature of the $i$-th clip.

\subsection{Group-aware Attention}
\label{subsec:goat}
There are two components of GOAT, that is, group-aware GCN and temporal-fusion attention. The overview of our framework is illustrated in Figure \ref{fig:goat}.

\textbf{Group-aware GCN.} This component is shown in the right part of Figure \ref{fig:goat}. We extract the group features by modeling the relations among actors with graphs. Given a video, we fix it into the given duration via downsampling or upsampling and divide it into several clips of equal length. Then we take three steps to extract the group features. First, we select the middle frame from each clip and build features for the actors. Specifically, we adopt object detection\cite{zhang2022dino} to get bounding boxes of actors. Then, we use Inception-V3\cite{szegedy2016rethinking} on the frame to get the feature map, to which we apply RoIAlign\cite{he2017mask} to extract features for each bounding box. We use a fully connected layer to get a $d$-dimensional vector from the feature map for each actor. Then we have a $N\times d$ representation denoted as $G$ for each middle frame, where $N$ denotes the number of actors and $d$ indicates the dimension of feature vectors.

Second, we build a relation graph with the feature matrix $G$ to model the relations among actors. Concretely, we regard each feature vector of an actor as a node in the graph. We follow the strategy in \cite{wu2019learning} to build edges among nodes by computing similarities of appearance features and relative location of actors.

Then, a Graph Convolutional Network (GCN)\cite{kipf2016semi} is adopted to the relation graph to enhance node features by weighted aggregation, which can be represented as:
\begin{equation}
    H^{(l+1)}=\sigma(AH^{(l)}W^{(l)}),
\end{equation}where $A$ indicates the adjacent matrix; $W^{(l)}$ is the weight matrix in $l$-th layer; $H^{(l)}$ denotes the features of nodes in $l$-th layer and $H^{(0)}=G$. Finally, The group feature vector is extracted by adopting average pooling to the sum of the input and the output features of GCN. The group feature vector of the $i$-th clip is denoted as $g_i$.

\textbf{Temporal-fusion Attention.} This component fuses the video embeddings along the temporal axis based on the representations from the group-aware GCN. As shown in the left part of Figure \ref{fig:goat}, for each clip split from the video, we send it into an I3D\cite{carreira2017quo} backbone as most previous works do\cite{xu2022finediving, yu2021group, tang2020uncertainty, zeng2020hybrid}. We leverage the strong capability of capturing global information of the multi-head attention encoder for learning group-aware video embeddings.

We denote the I3D features of the video as $\{\mathcal{F}(X_i)\}_{i=1}^T$, which serve as the ``\textit{value}" of the attention block. And the vectors $\{g_i\}_{i=1}^T$ from group-aware GCN serve as ``\textit{query}" and ``\textit{key}". The temporal-fusion attention module learns to discover the group feature correspondences among clips, and generates new temporal features in all of them. The module uses the relations of group features among different clips to assign different weights to different temporal features. The above attention learning can be represented as:
\begin{align}
    q^{(l)}&=W_Q^{(l)}q^{(l-1)},\\
    k^{(l)}&=W_K^{(l)}k^{(l-1)},\\
    W_{attn}^{(l)}&=Softmax(\frac{q^{(l)}{k^{(l)}}^T}{\sqrt{d}}),\\
    v^{(l)}&=BN[W_{attn}^{(l)}v^{(l-1)}+v^{(l-1)}],
\end{align}where $W_Q^{(l)},W_K^{(l)},W_{attn}^{(l)}$ indicate the learnable weights in $l$-th layer; $q^{(l)}, k^{(l)}, v^{(l)}$ denote the ``\textit{query}", ``\textit{key}" and ``\textit{value}" in $l$-th layer; $d$ is the dimension of feature vector and $BN$ is the BatchNorm block. Besides, $q^{(0)}=k^{(0)}=\{g_i\}_{i=1}^T, v^{(0)}=\{\mathcal{F}(X_i)\}_{i=1}^T$.

\subsection{Formation Detection}
\label{subsec:formationalgorithm}
In \ref{subsec:goat}, we propose a pipeline to fuse temporal representations based on a kind of spatial information, the group features. However, such a fusion strategy can also be based on other kinds of spatial features and even without any features by simply using self-attention. To prove the effectiveness of the group features, we devise another kind of spatial features, formation features, to make a comparison.

\textbf{Formation Detection Task.} As mentioned in Section \ref{sec:thelogodataset}, we use a polygon to depict the formation of actors. As shown in Figure \ref{fig:formations}, the polygon is represented by several coordinates of some actors but not all of them. So in this task, we need to distinguish whether an actor is the vertex of the polygon or not.

\textbf{Formation Detection Algorithm.} Inspired by the ``\textit{anchor}" in object detection\cite{ren2015faster}, we propose a feasible way to detect the vertexes.
Given a picture, we split it into 1024 patches of equal size. For each patch, we set the center point as the ``\textit{anchor}". We send the image into Inception-V3 and extract the $d$-dimensional feature vector with RoIAlign\cite{he2017mask} and a fully connected layer for each patch. To represent the absolute position of each patch, we use sine and cosine functions of different frequencies to encode the position of the ``\textit{anchor}". We add the $d$-dimensional positional encodings to the features and send them into a self-attention block to find the relations among different patches and build the $d$-dimensional formation features for them.

Then we take two steps to find vertexes. First, for each patch, we judge whether there is a vertex. Second, if there is a vertex, we calculate the relative coordinates of the vertex to the ``\textit{anchor}" point. So we detect vertexes by computing:
\begin{align}
    \hat{P}=[\hat{p}_1,...,\hat{p}_n], \quad  \hat{C}=[\hat{c}_1,...,\hat{c}_n],
\end{align}where $\hat{p}_i\in\mathbb{R}$ denotes the confidence of whether there is a vertex in the $i$-th patch and $\hat{c}_i\in\mathbb{R}^2$ is the relative coordinates of the vertex to the ``\textit{anchor}". In this way, the formation detection task is converted to a classification problem and a regression problem. And we respectively use a MLP to predict $\hat{P}$ and $\hat{C}$ upon the formation features mentioned above. The object function can be represented as:
\begin{gather}
    \mathcal{L}_{BCE}=-\sum_i[p_ilog\hat{p}_i+(1-p_i)log(1-\hat{p}_i)],\\
    \mathcal{L}_{MSE}=||C-\hat{C}||^2\centering.
\end{gather}

We minimize $\mathcal{L}_{BCE}$ and $\mathcal{L}_{MSE}$ to predict the vertexes.
The $d$-dimensional formation features of middle frame in each clip can replace the group features in \ref{subsec:goat} to fuse the temporal information. Relative experiments are conducted in our ablation study.

%% file: figures/goat.tex
\begin{figure*}[t]
    \centering
    \includegraphics[width=\linewidth]{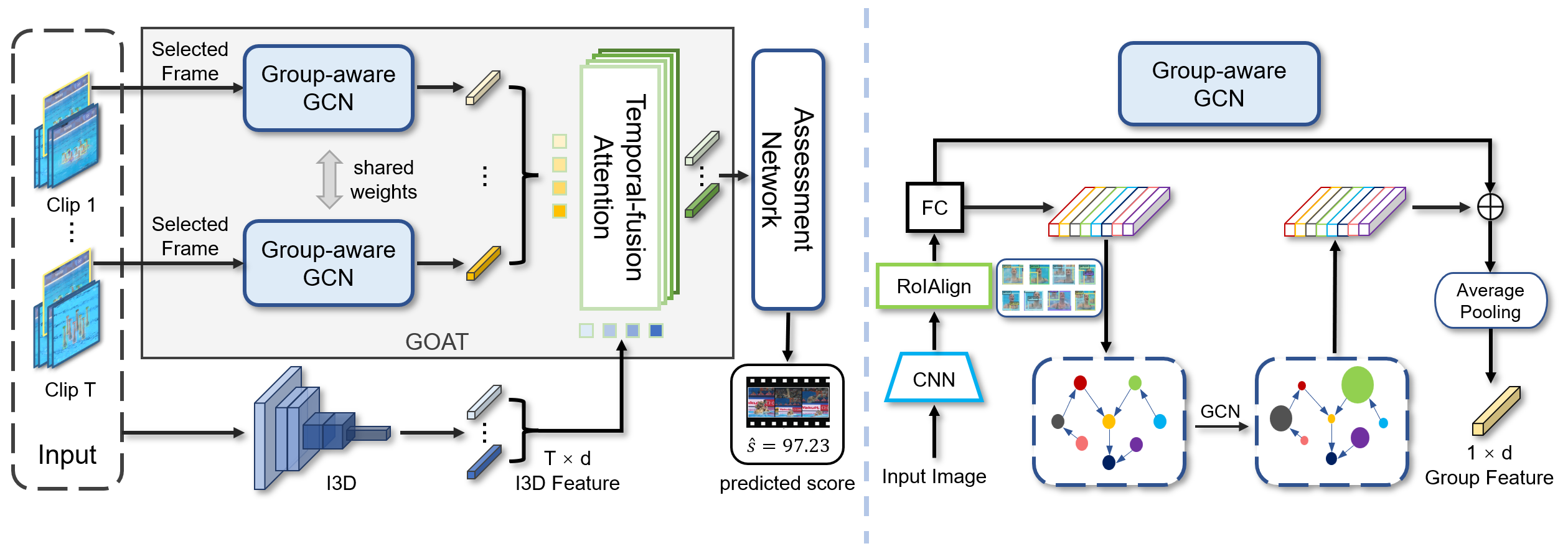}
    \vspace{-22pt}
    \caption{An overview of our group-aware approach for action quality assessment. First, we divide the video into several short clips of equal length. For each clip, we take the middle frame and perform object detection to get the bounding boxes of the actors. Then we send the frame into a CNN to extract the features of the actors. Then we use the feature vector of each actor as the node to construct the relation graph. We use Graph Convolutional Network to enhance the features in the graph, and send the output features into GOAT as ``\textit{queries}" and ``\textit{keys}". And the ``\textit{values}" are the features obtained from the clip by the video feature backbone such as I3D and video swin-transformer. Thus the aggregation in time can be completed. Finally, the output features are sent into the assessment network to predict the scores.}
    \label{fig:goat}
    \vspace{-10pt}
\end{figure*}

%% file: sections/5_experiments.tex
\section{Experiments}
\label{sec:experiment}
\subsection{Implementation Details}

\textbf{Action Quality Assessment.} Following \cite{zeng2020hybrid, parmar2017learning, xu2019learning}, we take two stages to assess action quality: feature extraction and score prediction. In the feature extraction stage, we use the I3D model pretrained on Kinetics\cite{carreira2017quo} dataset to extract the 1024-dimensional feature vector for each video clip. Specifically, we sample 5406 frames for each video, split them into 540 snippets, and fed them into I3D. Each snippet contains 16 continuous frames with a stride of 10 frames. In the ablation study, we also use the video swin-transformer\cite{liu2022video} as the backbone, which is also pretrained on Kinetics. We follow the sampling strategy in \cite{liu2022video} to use a temporal stride of 2. The Inception-V3 in \ref{subsec:goat} is pretrained on ImageNet\cite{deng2009imagenet} to extract the CNN features. 
Following \cite{xu2022finediving, tang2020uncertainty, yu2021group}, we split the dataset into 3:1 for training and testing in all the experiments. We also test our model using the Mindspore\cite{Mindspore}. 

\textbf{Action Segmentation.} We conduct action segmentation on LOGO with several existing approaches (ASFormer\cite{yi2021asformer}, SSTDA\cite{9186840}, MS\_TCN++\cite{chen2020action}, ASRF\cite{ishikawa2021alleviating})to provide a benchmark. We try two backbones to extract frame-wise features: I3D\cite{carreira2017quo} model and video swin-transformer\cite{liu2022video} pretrained on Kinetics. 
For SWIN, we expand the snippet size to 32 frames to extract 1536-dimensional features. In the training stage, we train action segmentation approaches on their default setting. 
\input{figures/attentionvisualization}
\subsection{Evaluation Metrics}

\textbf{Action Quality Assessment.} Following \cite{xu2022finediving, yu2021group, tang2020uncertainty, zeng2020hybrid}, we comprehensively evaluate our approach to AQA under two metrics, Spearman's rank correlation ($\rho$) and relative $\ell_2$-distance (R-$\ell_2$). Spearman's rank correlation is used to evaluate the rank correlation between the prediction and ground-truth while the relative $\ell_2$-distance (R-$\ell_2$) focuses on the difference of the numerical values.

\textbf{Action Segmentation.} Following\cite{yi2021asformer,9186840,chen2020action,ibrahim2018hierarchical}, we adopt three widely used evaluation metrics including frame-wise accuracy (Acc), segmental edit distance and segmental F1 score at overlapping thresholds 10\%,  25\%, and 50\%, denoted by F1@\{10, 25, 50\} to evaluate the performance of action segmentation approaches.

\subsection{Comparison with the State of the Art}

\textbf{Action Quality Assessment.} Table \ref{table:aqa_experiment_result} shows the experimental results, which reveal great challenges to performing action quality assessment on the LOGO dataset. With the I3D backbone, the mainstream methods in single-person short-term AQA\cite{xu2022finediving, tang2020uncertainty, yu2021group} attain the result of 0.4259, 0.4712, and 0.4518 on Spearman's rank correlation respectively. Besides, the results show that our approach achieves the state-of-the-art. We observe that USDL+GOAT, CoRe+GOAT, and TSA+GOAT consistently improve the performance over the original models. Specifically, our approach respectively obtained 8.48\%, 4.73\%, 
and 10.12\% improvements on Spearman's rank correlation. Our method also outperforms the previous long-term AQA method\cite{zeng2020hybrid}. The experimental results illustrate the effectiveness of our proposed module when dealing with multi-person long-term scenarios. 

\input{tables/experiment_aqa_comparison}
\input{tables/action_segmentation}

\textbf{Action Segmentation.} Table \ref{table:gtea1} presents the experimental results. With the I3D feature, the previous action segmentation methods \cite{yi2021asformer,9186840,chen2020action,ishikawa2021alleviating} achieve the result of 73.9\%, 70.6\%, 63.4\% and 69.8\% frame accuracy respectively. And given an identical setting, we see that the SWIN-pretrained feature obtained significant improvement in all five metrics. Table \ref{table:asgoat} reveals that our method could improve the performance of MS\_TCN++ and ASFormer and achieve the state-of-the-art with the group information. 

\subsection{Generalization of GOAT}
\input{tables/generalization}
Our method is currently suitable for multi-person scenarios. We select current AQA datasets with two-player scenes (there are only 2 players at most for now) and redivide them according to the number of people. TASD-2\cite{gao2020asymmetric} is a dataset collected from synchronized diving events with 2 players in each video. Based on the divided dataset, we obtain results in Table \ref{table:generalization}, which prove that GOAT can improve the AQA results of two-player scenes in other datasets. These results also demonstrate the generalization of GOAT. Our method can perform well in short-term, two-player scenes.

\input{figures/formationvisualization.tex}
\subsection{Ablation Study}

\input{tables/ablation}
\input{tables/action_segmentation_goat.tex}
We change the video feature backbone to verify the generality of our approach. Besides, we also try to use other spatial features or simply use self-attention for temporal information fusion to demonstrate the effectiveness of the group features. The experiments also illustrate that our temporal fusion strategy outperforms the average pooling.

\textbf{Spatial Features.} As shown in Table \ref{table:ablation}, we conduct experiments utilizing formation features to replace the group embeddings used in GOAT to fuse the temporal representations. The formation features are extracted from the attention block output in the formation detection pipeline mentioned in \ref{subsec:formationalgorithm}, which is pretrained on the LOGO dataset by executing formation detection. We also adopt self-attention to replace GOAT, which means we perform our temporal fusion strategy without the assistance of any spatial information.
The experimental results show that the formation features outperform the original models and self-attention models but perform worse than the group features in most cases. The results also show that the self-attention models perform better than the original models. Such results of our experiments prove that: (1) Our temporal fusion strategy outperforms the average pooling even without any spatial information. (2) By modeling the spatial information, the performance of our proposed approach is improved. (3) Building the relations among actors, the group features perform better than the formation features.

\textbf{Video Feature Backbone.} As shown in Table \ref{table:ablation} and Table \ref{table:aqa_experiment_result}. We change the backbone to video swin-transformer (abbreviated as SWIN) and repeat all our experiments. Similar to the results shown in the action segmentation experiments, the methods based on SWIN perform much better than the methods based on I3D in most cases, which illustrates the effectiveness of SWIN. Besides, except for the case of TSA+Self, the results of our methods substantially outperform the original models based on SWIN, which proves the generality of our approach. 

\textbf{Asist AQA with Action Segmentation.} To assist the AQA task with the action segmentation task (or action labels), we use the backbone features finetuned in the action segmentation task as the input of GOAT in the experiments shown in the lower part of Table \ref{table:ablation}. The results illustrate that with the help of the segmentation task, the performances of AQA are improved compared with the original methods.

\subsection{Visualization}
We visualize the output of GOAT, as shown in Figure \ref{fig:attentionvisualization}. It shows that GOAT highlights the part where actors perform effective action with clear formations, demonstrating the effectiveness of our temporal fusion strategy. We also visualize the prediction results of our formation detection module, as shown in Figure \ref{fig:formationvisualization}. It can be seen that our approach can distinguish the vertexes of the formations.

%% file: figures/attentionvisualization.tex
\begin{figure*}[t]
    \centering
    \includegraphics[width=\linewidth]{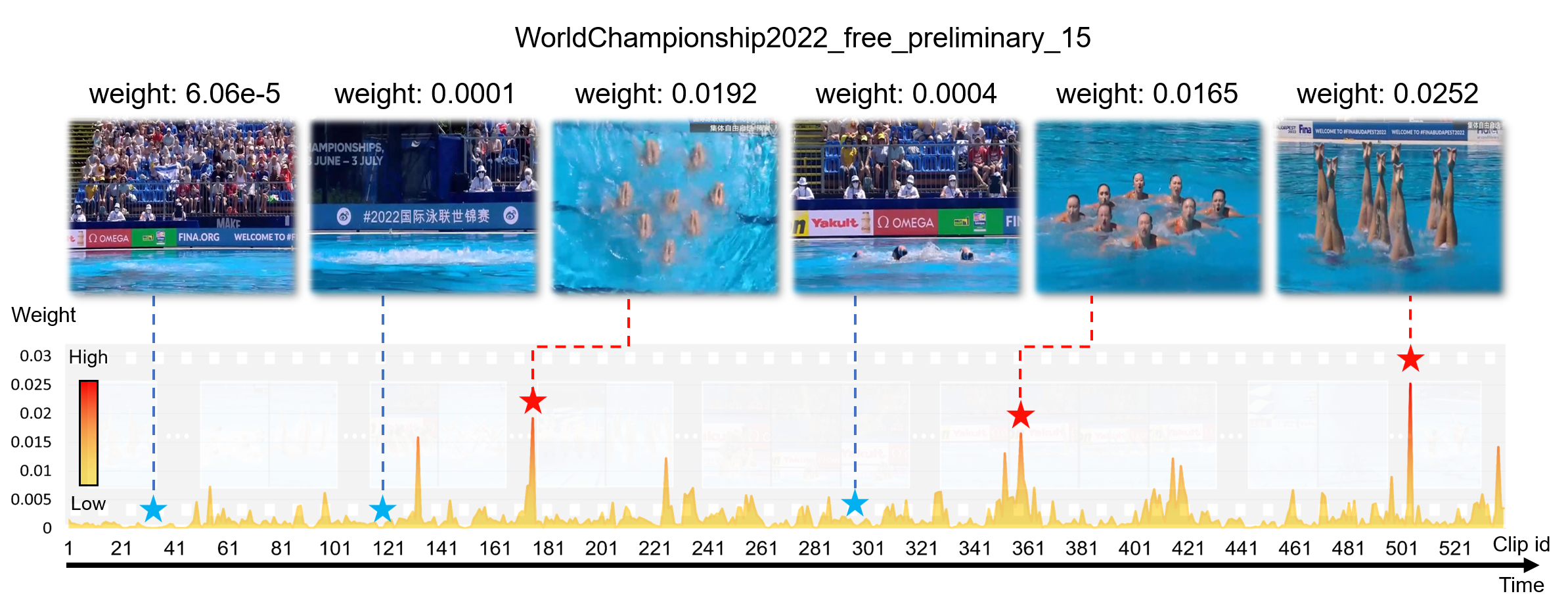}
    \vspace{-20pt}
    \caption{The visualization of the output of our proposed GOAT in action quality assessment. We use red stars to denote clips with high weight while using blue stars to represent clips with low weight. Our approach can focus on where the athletes perform effective movements with clear formations while it can also ignore the redundant part such as all actors are under-water.}
    \label{fig:attentionvisualization}
    \vspace{-12pt}
\end{figure*}

%% file: tables/experiment_aqa_comparison.tex
\begin{table}[t]
\caption{Comparisons of performance with existing AQA methods on LOGO. The higher $\rho$, the lower R-$\ell_2$, the better performance.}
\vspace{-5pt}
\footnotesize
\renewcommand\tabcolsep{1.2mm}
\label{table:aqa_experiment_result}
\centering
\begin{tabular}{l|cc|cc}
\toprule
\multirow{2}{*}{Method} &\multicolumn{2}{c|}{I3D} & \multicolumn{2}{c}{SWIN}  \\
\cline{2-5}
& $\rho$ $\uparrow$ & R-$\ell_2(\times100)$ $\downarrow$ & $\rho$ $\uparrow$ & R-$\ell_2(\times100)$ $\downarrow$\\
\midrule
USDL\cite{tang2020uncertainty} & 0.4259 & 5.7364 & 0.4725 & 5.0762 \\
CoRe\cite{yu2021group} & 0.4712 & 5.4086 & 0.5002 & 5.9597 \\
TSA\cite{xu2022finediving} & 0.4518 & 5.5326 & 0.4751 & 4.7778 \\
ACTION-NET\cite{zeng2020hybrid} & 0.3057 & 5.8581 & 0.4101 & 5.5693 \\
\midrule
\textbf{USDL\cite{tang2020uncertainty}+GOAT} & 0.4620 & \textbf{4.8739} & 0.5349 & 5.0220 \\
\textbf{CoRe\cite{yu2021group}+GOAT} & \textbf{0.4935} & 5.0716 & \textbf{0.5599} & \textbf{4.7626} \\
\textbf{TSA\cite{xu2022finediving}+GOAT} & 0.4855 & 5.3943 & 0.4843 & 5.4086 \\
\bottomrule
\end{tabular}
\vspace{-10pt}
\end{table}

%% file: tables/action_segmentation.tex
\begin{table}[]
\small
\begin{center}
\setlength{\tabcolsep}{5pt}
\caption{Action segmentation results on LOGO.}
\vspace{-5pt}
 \setlength{\tabcolsep}{1.57mm}{
\begin{tabular}{l|l|ccc|c|c}
\toprule
Method & Features & \multicolumn{3}{c|}{F1@\{10,25,50\}} & Edit & Acc \\ \midrule
ASFormer\cite{yi2021asformer} & SWIN & 81.7 & 79.8 & 71.1 & 75.3 & 81.0 \\
ASFormer\cite{yi2021asformer} & I3D & 75.1 & 71.6 & 61.3 & 68.2 & 73.9 \\
MS\_TCN++\cite{9186840} & SWIN & 80.8 & 78.8 & 70.2 & 73.5 & 80.1 \\
MS\_TCN++\cite{9186840} & I3D & 72.6 & 69.2 & 58.5 & 66.0 & 70.6 \\
SSTDA\cite{chen2020action} & SWIN & 77.8 & 75.9 & 66.3 & 70.0 & 79.2 \\
SSTDA\cite{chen2020action} & I3D & 60.1 & 55.7 & 44.0 & 50.1 & 63.4 \\
ASRF\cite{ishikawa2021alleviating} & SWIN & 80.8 & 79.1 & 72.1 & 73.2 & 80.0 \\
ASRF\cite{ishikawa2021alleviating} & I3D & 73.6 & 70.5 & 59.8 & 66.8 & 69.8 \\
\bottomrule
\end{tabular}}\label{table:gtea1}
\vspace{-25pt}
\end{center}
\end{table}

%% file: tables/generalization.tex


\begin{table}[t]
\caption{Comparisons of AQA results of CORE and CORE+GOAT based on existing short-term, two-player datasets. \textit{@2} means AQA datasets with two-player scenes.}
\vspace{-5pt}
\footnotesize
\renewcommand\tabcolsep{1.2mm}
\label{table:generalization}
\centering
\begin{tabular}{l|cc|cc}
\toprule
\multirow{2}{*}{Dataset} &\multicolumn{2}{c|}{CORE\cite{yu2021group}} & \multicolumn{2}{c}{CORE\cite{yu2021group}+GOAT} \\
\cline{2-5}
& $\rho$ $\uparrow$ & R-$\ell_2(\times100)$ $\downarrow$ & $\rho$ $\uparrow$ & R-$\ell_2(\times100)$ $\downarrow$ \\
\midrule
FineDiving\cite{xu2022finediving}@2 & 0.8991 & 0.3751 & 0.9032 & 0.3529\\
TASD-2\cite{gao2020asymmetric}@2 & 0.9189 & 0.7863 & 0.9334 & 0.7054\\
AQA-7\cite{parmar2019action}@2 & 0.9012 & 0.7302 & 0.9325 & 0.6423\\
\bottomrule
\end{tabular}
\vspace{-10pt}
\end{table}

%% file: figures/formationvisualization.tex
\begin{figure*}[t]
    \centering
    \includegraphics[width=\linewidth]{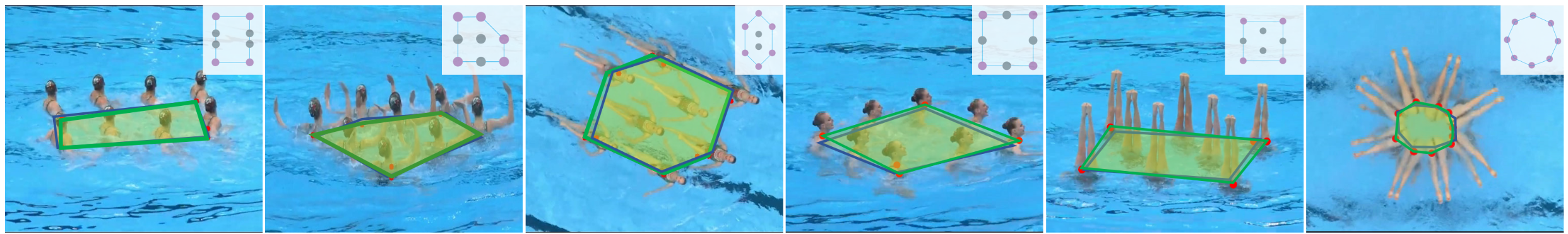}
    \vspace{-15pt}
    \caption{The visualization of the prediction results of our formation detector module. The green polygons represent prediction results and the yellow polygons with blue edges are the ground truth. The results show that our approach can detect the positions of actors and distinguish whether the athlete is the formation vertex or not, which guarantees the reliability of the formation features.}
    \label{fig:formationvisualization}
    \vspace{-10pt}
\end{figure*}

%% file: tables/ablation.tex
\begin{table}[t]
\caption{Ablation studies on LOGO. \textit{Self.} indicates self-attention; \textit{Form.} indicates using formation features for temporal fusing; \textit{AS.} means using the action segmentation features for temporal fusing.}
\vspace{-5pt}
\footnotesize
\renewcommand\tabcolsep{1.3mm}
\label{table:ablation}
\centering
\begin{tabular}{l|cc|cc}
\toprule
\multirow{2}{*}{Method} &\multicolumn{2}{c|}{I3D} & \multicolumn{2}{c}{SWIN}  \\
\cline{2-5}
& $\rho$ $\uparrow$ & R-$\ell_2(\times100)$ $\downarrow$ & $\rho$ $\uparrow$ & R-$\ell_2(\times100)$ $\downarrow$\\
\midrule
USDL\cite{tang2020uncertainty}+Self. & 0.4335 & 4.8898 & 0.5078 & 4.1305 \\
CoRe\cite{yu2021group}+Self. & 0.4348 & 5.6019 & 0.5269 & 4.5072 \\
TSA\cite{xu2022finediving}+Self. & 0.4585 & 5.0336 & 0.4720 & 5.7280 \\
\midrule
USDL\cite{tang2020uncertainty}+Form. & 0.4355 & 5.7636 & 0.5370 & 5.0888 \\
CoRe\cite{yu2021group}+Form. & 0.4820 & 5.3003 & 0.5401 & 5.1747 \\
TSA\cite{xu2022finediving}+Form. & 0.4241 & 5.1821 & 0.4903 & 4.8627 \\
\midrule
USDL\cite{tang2020uncertainty}+AS. & 0.4512 & 4.8837 & 0.5108 & 5.0479 \\
CoRe\cite{yu2021group}+AS. & 0.4825 & 5.2393 & 0.5364 & 4.9102 \\
TSA\cite{xu2022finediving}+AS. & 0.4772 & 5.4203 & 0.4829 & 5.3561 \\
\bottomrule
\end{tabular}
\vspace{-10pt}
\end{table}

%% file: tables/action_segmentation_goat.tex
\begin{table}[]
\small
\begin{center}
\caption{The action segmentation results of existing methods with GOAT on LOGO. \textit{+GOAT} indicates the method with GOAT.}
\vspace{-5pt}
\setlength{\tabcolsep}{0.9mm}{
\begin{tabular}{l|l|ccc|c|c}
\toprule
Method & Features & \multicolumn{3}{c|}{F1@\{10,25,50\}} & Edit & Acc \\ 
\midrule
MS\_TCN++\cite{9186840}+GOAT & SWIN & 80.9 & 79.0 & 70.5 & 74.0 & 81.1 \\
MS\_TCN++\cite{9186840}+GOAT & I3D & 73.2 & 69.9 & 59.4 & 67.3 & 71.7 \\
ASFormer\cite{yi2021asformer}+GOAT & SWIN & 82.2 & 80.3 & 73.6 & 75.9 & 81.7 \\
ASFormer\cite{yi2021asformer}+GOAT & I3D & 75.8 & 72.0 & 62.1 & 68.8 & 74.6 \\
\bottomrule
\end{tabular}}\label{table:asgoat}
\vspace{-15pt}
\end{center}
\end{table}

%% file: sections/6_conclusion.tex
\section{Conclusion}
In this paper, we construct the first multi-person long-form video dataset, LOGO, for action quality assessment. We also propose a group-aware module, GOAT, to build relations among multiple actors and fuse the temporal representations based on spatial information. Furthermore, the utilization of GOAT in action quality assessment and action segmentation both achieve substantial improvements compared to the existing methods.


\textbf{Existing Assets and Personal Data.} The videos in LOGO are downloaded from several websites such as YouTube. We are actively connecting with the authors to ensure that appropriate consent has been obtained.

\textbf{Acknowledgments.} This work was sponsored in part by the National Natural Science Foundation of China (Grant No. 62206153, 62125603), CAAI-Huawei MindSpore Open Fund, Deng Feng Fund, Young Elite Scientists Sponsorship Program by CAST (No. 2022QNRC001), and Shenzhen Stable Supporting Program (WDZC20220818112518001).